\documentclass{article}
\usepackage{spconf,amsmath,graphicx}
\usepackage{hyperref, xspace}
\usepackage{tikz}
\usepackage{booktabs}

\tikzset{every picture/.style={line width=0.75pt}} 

\newcommand{\parenthesis}[1]{\left(#1\right)}
\newcommand{\brackets}[1]{\left[#1\right]}

\newcommand{\at}[1][t]{\ensuremath{\gls*{action}_{#1}}\xspace}
\newcommand{\st}[1][t]{\ensuremath{\gls*{state}_{#1}}\xspace}
\newcommand{\phit}[1][t]{\ensuremath{\gls*{feat}\parenthesis{\st[#1]}}\xspace}

\newcommand{\advt}[1][t]{\ensuremath{\gls*{advantage}_{\gls*{policy}}\parenthesis{\st[#1], \at[#1]}} \xspace}

\newcommand{\stsp}[1][]{\ensuremath{\gls*{state_space}^{#1}} \xspace}

\usepackage[acronym]{glossaries}

\newacronym{ai}{AI}{Artificial Intelligence}
\newacronym{ml}{ML}{Machine Learning}
\newacronym{dl}{DL}{Deep Learning}
\newacronym{gan}{GAN}{Generative Adversarial Network}

\newacronym{rl}{RL}{Reinforcement Learning}
\newacronym{mdp}{MDP}{Markov Decision Process}
\newacronym{dqn}{DQN}{Deep Q Network}
\newacronym{trpo}{TRPO}{Trust Region Policy Optimization}
\newacronym{ppo}{PPO}{Proximal Policy Optimization}
\newacronym{ac}{AC}{Actor-Critic}
\newacronym{a2c}{A2C}{Advantage Actor-Critic}
\newacronym{a3c}{A3C}{Asynchronous Advantage Actor-Critic}
\newacronym{atta2c}{AttA2C}{Attention-aided Advantage Actor-Critic}
\newacronym{acer}{ACER}{Actor-Critic with Experience Replay}
\newacronym{icm}{ICM}{Intrinsic Curiosity Module}
\newacronym{rcm}{RCM}{Rational Curiosity Module}

\newacronym{dnn}{DNN}{Deep Neural Network}
\newacronym{nn}{NN}{Neural Network}
\newacronym{ann}{ANN}{Artificial Neural Network}
\newacronym{rnn}{RNN}{Recurrent Neural Network}
\newacronym{fc}{fc}{fully connected layer}
\newacronym{cn}{conv}{Convolutional layer}
\newacronym{cnn}{CNN}{Convolutional Neural Network}
\newacronym{vae}{VAE}{Variational Autoencoder}

\newacronym{lstm}{LSTM}{Long Short-Term Memory}
\newacronym{gru}{GRU}{Gated Recurrent Unit}
\newacronym{relu}{ReLU}{Rectified Linear Unit}
\newacronym{bn}{BN}{Batch Normalization}

\newacronym{sgd}{SGD}{Stochastic Gradient Descent}
\newacronym{adam}{ADAM}{Adaptive Moment Estimation}
\newacronym{pg}{PG}{Policy Gradient}

\newacronym{gpu}{GPU}{Graphics Processing Unit}
\newacronym{mse}{MSE}{Mean Squared Error}
\newacronym{mae}{MAE}{Mean Absolute Error}

\newglossary{abbrev}{abs}{abo}{Nomeclature}
\newglossaryentry{policy}{
    name        = \ensuremath{\pi} ,
    description = {policy of an agent} ,
    type        = abbrev,
}

\newglossaryentry{action}{
    name        = \ensuremath{a} ,
    description = {action of an agent} ,
    type        = abbrev,
}

\newglossaryentry{action_space}{
    name        = \ensuremath{\mathcal{A}} ,
    description = {action space of an agent} ,
    type        = abbrev,
}
\newglossaryentry{state}{
    name        = \ensuremath{s} ,
    description = {state of an agent} ,
    type        = abbrev,
}

\newglossaryentry{state_space}{
    name        = \ensuremath{\mathcal{S}} ,
    description = {state space of an agent} ,
    type        = abbrev,
}

\newglossaryentry{feat}{
    name        = \ensuremath{\phi} ,
    description = {feature transform} ,
    type        = abbrev,
}

\newglossaryentry{state_trans_prob_mat}{
    name        = \ensuremath{\mathcal{P}} ,
    description = {state transition probability matrix} ,
    type        = abbrev,
}

\newglossaryentry{state_trans_prob}{
    name        = \ensuremath{p} ,
    description = {state transition probability} ,
    type        = abbrev,
}

\newglossaryentry{reward_set}{
    name        = \ensuremath{\mathcal{R}} ,
    description = {set of the rewards in an environment} ,
    type        = abbrev,
}

\newglossaryentry{reward}{
    name        = \ensuremath{r} ,
    description = {reward in an environment} ,
    type        = abbrev,
}

\newglossaryentry{return}{
    name        = \ensuremath{g} ,
    description = {return in an environment} ,
    type        = abbrev,
}

\newglossaryentry{discount_factor}{
    name        = \ensuremath{\gamma} ,
    description = {discount factor of the reward} ,
    type        = abbrev,
}

\newglossaryentry{q_func}{
    name        = \ensuremath{Q} ,
    description = {action-value function} ,
    type        = abbrev,
}

\newglossaryentry{state_value_func}{
    name        = \ensuremath{V} ,
    description = {state-value function} ,
    type        = abbrev,
}

\newglossaryentry{advantage}{
    name        = \ensuremath{A} ,
    description = {advantage function} ,
    type        = abbrev,
}

\newglossaryentry{nn_param}{
    name        = \ensuremath{\theta} ,
    description = {parameters of a \acrshort{dnn}} ,
    type        = abbrev,
}

\newglossaryentry{cost_fn}{
    name        = \ensuremath{J},
    description = {cost function of the optimization problem} ,
    type        = abbrev,
}

\newglossaryentry{cost_fwd}{
    name        = \ensuremath{J_{fwd}},
    description = {cost function of the \acrshort{icm} forward model} ,
    type        = abbrev,
}

\newglossaryentry{cost_inv}{
    name        = \ensuremath{J_{inv}},
    description = {cost function of the \acrshort{icm} inverse model} ,
    type        = abbrev,
}

\newglossaryentry{cost_icm}{
    name        = \ensuremath{J_{icm}},
    description = {cost function of the \acrshort{icm} model} ,
    type        = abbrev,
}

\newglossaryentry{cost_rcm}{
    name        = \ensuremath{J_{rcm}},
    description = {cost function of the \acrshort{rcm} model} ,
    type        = abbrev,
}

\newglossaryentry{cost_a2c}{
    name        = \ensuremath{J_{A2C}},
    description = {cost function of the \acrshort{a2c} model} ,
    type        = abbrev,
}

\newglossaryentry{st_traj}{
    name        = \ensuremath{\tau},
    description = {state trajectory} ,
    type        = abbrev,
}
\newglossaryentry{pred}{
    name        = \ensuremath{y} ,
    description = {output value of a \acrshort{dnn}} ,
    type        = abbrev,
}
\newglossaryentry{input}{
    name        = \ensuremath{\mathbf{x}} ,
    description = {input vector of a layer} ,
    type        = abbrev,
}

\newglossaryentry{icm_beta}{
    name        = \ensuremath{\beta} ,
    description = {weighting factor of \gls{cost_fwd} and \gls{cost_inv} in \gls{icm}} ,
    type        = abbrev,
}

\newglossaryentry{lr}{
    name        = \ensuremath{\eta} ,
    description = {parameter of \acrshort{sgd} which defines the rate of change of the weight matrices} ,
    type        = abbrev,
}
\newglossaryentry{weight_mat}{
    name        = \ensuremath{\mathbf{W}},
    description = {weight matrix of a \acrshort{dnn}} ,
    type        = abbrev,
}

\newglossaryentry{weight}{
    name        = \ensuremath{w_i} ,
    description = {element of a \gls{weight_mat}} ,
    type        = abbrev,
}

\newglossaryentry{bias}{
    name        = \ensuremath{\mathbf{b}} ,
    description = {bias vector of a layer} ,
    type        = abbrev,
}
\newglossaryentry{act}{
    name        = \ensuremath{f} ,
    description = {activation function of a layer} ,
    type        = abbrev,
}
\newglossaryentry{act_out}{
    name        = \ensuremath{\mathbf{x}_{act}} ,
    description = {activated output of a layer} ,
    type        = abbrev,
}
\newglossaryentry{mean}{
    name        = \ensuremath{\mu} ,
    description = {mean of a dataset} ,
    type        = abbrev,
}
\newglossaryentry{std}{
    name        = \ensuremath{\sigma} ,
    description = {standard deviation of a dataset} ,
    type        = abbrev,
}

\newglossaryentry{mom_c}{
    name        = \ensuremath{\mu_m},
    description = {weighting factor in the case of momentum methods} ,
    type        = abbrev,
}

\newglossaryentry{bn_a}{
    name        = \ensuremath{\alpha_{BN}},
    description = {linear coefficient of \acrshort{bn}} ,
    type        = abbrev,
}
\newglossaryentry{bn_b}{
    name        = \ensuremath{\beta_{BN}},
    description = {bias coefficient of \acrshort{bn}} ,
    type        = abbrev,
}
\newglossaryentry{eps}{
    name        = \ensuremath{\epsilon},
    description = {small constant for maintaining numeric stability} ,
    type        = abbrev,
}
\newglossaryentry{hadamard}{
    name        = \ensuremath{\odot},
    description = {Hadamard-product, used for the notation of element-wise product for matrices} ,
    type        = abbrev,
}

\newglossaryentry{p}{
    name        = \ensuremath{p},
    description = {probability} ,
    type        = abbrev,
}

\newglossaryentry{layers}{
    name        = \ensuremath{\mathcal{L}},
    description = {the set of affected layers in \acrshort{wmm}-methods} ,
    type        = abbrev,
}

\newglossaryentry{layer_c}{
    name        = \ensuremath{\mathcal{L}_c},
    description = {the set of affected layers components in \acrshort{wmm}-methods, such as the gates of \acrshort{lstm}} ,
    type        = abbrev,
}

\newglossaryentry{coverage}{
    name        = \ensuremath{c},
    description = {the ratio of a layer's dimension which can be affected in \acrshort{wmm}-methods} ,
    type        = abbrev,
}

\newcommand{\ie}{i.e.\@\xspace}
\newcommand{\Ie}{I.e.\@\xspace}
\newcommand{\eg}{e.g.\@\xspace}
\newcommand{\Eg}{E.g.\@\xspace}

\newcommand{\wrt}{w.r.t.\@\xspace} 

\title{Attention-based Curiosity-driven Exploration in
Deep Reinforcement Learning}
\name{Patrik Reizinger\sthanks{Supported by the \'UNKP-19-2 New National Excellence Program of the Ministry for Innovation and Technology} and M\'arton Szemenyei}
\address{Budapest University of Technology and Economics\\ Department of Control Engineering and Information Technology\\ Budapest, Hungary\\ rpatrik96@sch.bme.hu}
\begin{document}
%
\maketitle
\begin{abstract}
\acrlong*{rl} enables to train an agent via interaction with the environment. However, in the majority of real-world scenarios, the extrinsic feedback is sparse or not sufficient, thus intrinsic reward formulations are needed to successfully train the agent. This work investigates and extends the paradigm of curiosity-driven exploration. First, a probabilistic approach is taken to exploit the advantages of the attention mechanism, which is successfully applied in other domains of \acrlong*{dl}. Combining them, we propose new methods, such as \acrshort*{atta2c}, an extension of the \acrlong*{ac} framework. Second, another curiosity-based approach - \acrshort*{icm} - is extended. The proposed model utilizes attention to emphasize features for the dynamic models within \acrshort*{icm}, moreover, we also modify the loss function, resulting in a new curiosity formulation, which we call rational curiosity. The corresponding implementation can be found at \url{https://github.com/rpatrik96/AttA2C/}.
\end{abstract}
%

\section{Introduction}
\label{sec:intro}
Learning through reinforcement is a powerful concept, as it - similarly to unsupervised learning - 
does not need labeled samples, which are both time- and money-consuming to collect.
Moreover, \gls*{rl} is rather similar to the concept of human learning, since it defines the learning process as an interaction between an agent and its environment. Interestingly, in some cases, it was proven that an \gls*{rl} agent optimizes the same objective function as primates do, as noted in ~\cite{sutton}.

Speaking of optimization, an optimum can only be referred to \wrt some given 
criteria, thus \gls*{rl} has to define its own criterion, as it is the case for \gls*{ai} generally.
In optimization, we generally face the so-called bias-variance problem, which
has a new designation in the \gls*{rl} domain: it is called the exploration-exploitation dilemma. Here, exploitation refers to preferring prior knowledge, \ie bias. In some sense, exploitation can be thought of as a risk-minimizing way of actions. Nonetheless, this policy may not always be the best option. Clearly, some sort of compromise should be obtained between exploitation and exploration - which stands for a variance-preferring policy.

\gls*{rl} agents are generally trained based on the feedback - rewards - collected from the environment. However, the environment is not guaranteed to have an inner dynamics that rewards the agent in a way that corresponds with its goals, \eg if the rewards are not negative the agent may collect an infinite amount of reward given an infinite time horizon, while it can fail its objective if it is to finish the task as quickly as possible. Thus, only having an extrinsic incentive may be impractical. Furthermore, rewards are in most real-life scenarios sparse, so it is more difficult for the agent to 
assign credit to useful actions and omit disadvantageous ones in the future.

To mitigate the above-mentioned conflict, several methods were developed to supplement the objective function of an \gls*{rl} agent to improve its performance. On the higher level, \ie considering model-based \gls*{rl}, the World model architecture~\cite{world_models} is one answer for these challenges.
Curiosity-based methods, which are the main topic of this work, modify the loss function (or even the network architecture) by adding terms to incentivize exploration. In this topic, the \gls*{icm} module of \cite{icm} is a major contribution, which will be used extensively in this work to build upon. \gls*{icm} introduces inner dynamics (forward and inverse) to quantify the prediction error of the next state and action, which is utilized as an intrinsic reward - a more extensive study can be found in \cite{large_scale_curiosity}. Another promising work in this field is the notion of disagreement, which is an ensemble-based curiosity formulation~\cite{disagreement}. 

In this paper, we explore the capabilities of applying attention~\cite{attention} in order to incentivize exploration and improve generalization. We test our proposed methods using Atari games from OpenAI Gym~\cite{gym}.
Our main contributions are as follows:
\begin{itemize}
    \itemsep-.55em 
    \item \acrshort*{atta2c}, an attention-aided \acrshort*{a2c} variant,
    \item feature- and action-selective extension of the \gls*{icm}~\cite{icm},
    \item a rational curiosity formulation.
\end{itemize}


\section{Attention-based Curiosity}
\label{sec:attn_curiosity}
After reviewing the approaches of the literature, we turn to the main contributions of this work, which are based on the combination of the advancements of different \gls*{dl} fields, mainly manifesting in the introduction of the attention mechanism to curiosity-driven \gls*{rl} - for use in other domains, see \eg \cite{attn_gan}.
Using attention, \gls*{a2c}~\cite{a3c} (one of the standard architectures of \gls*{rl}) and the module responsible for curiosity-driven exploration of ~\cite{icm}, modifications are carried out to include prior assumptions in the model.

\subsection{AttA2C}
\label{subsec:atta2c}
The main idea for \gls*{atta2c} comes from the \gls*{gan}\cite{gan}--\acrlong*{ac} correspondence~\cite{ac_gan}, which is summarized in \autoref{tab:ac_gan}. Highlighting the only difference (typeset in bold in the table), \ie the fact that while an \acrlong*{ac} network feeds the same input/features to its both heads, the \gls*{gan} does quite the opposite. 

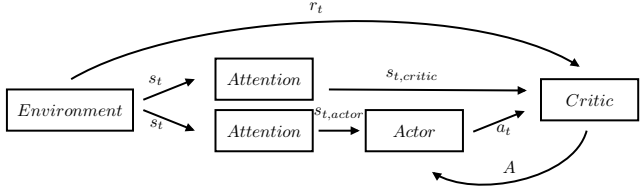
\begin{figure}[tb]
    \centering
    \begin{tikzpicture}[x=0.75pt,y=0.75pt,yscale=-.8,xscale=.8, scale=0.65, every node/.style={scale=0.65}]

\draw    (320,230) -- (358,230) ;
\draw [shift={(360,230)}, rotate = 180] [fill={rgb, 255:red, 0; green, 0; blue, 0 }  ][line width=0.75]  [draw opacity=0] (8.93,-4.29) -- (0,0) -- (8.93,4.29) -- cycle    ;

\draw    (330,190) -- (520,190.99) ;
\draw [shift={(522,191)}, rotate = 180.3] [fill={rgb, 255:red, 0; green, 0; blue, 0 }  ][line width=0.75]  [draw opacity=0] (8.93,-4.29) -- (0,0) -- (8.93,4.29) -- cycle    ;

\draw    (518.14,211.74) -- (470,231) ;

\draw [shift={(520,211)}, rotate = 158.2] [fill={rgb, 255:red, 0; green, 0; blue, 0 }  ][line width=0.75]  [draw opacity=0] (8.93,-4.29) -- (0,0) -- (8.93,4.29) -- cycle    ;
\draw    (431.77,271.19) .. controls (471.44,296.64) and (567.2,277.28) .. (580,230) ;

\draw [shift={(430,270)}, rotate = 35.26] [fill={rgb, 255:red, 0; green, 0; blue, 0 }  ][line width=0.75]  [draw opacity=0] (8.93,-4.29) -- (0,0) -- (8.93,4.29) -- cycle    ;
\draw   (17.7,190) -- (140,190) -- (140,230) -- (17.7,230) -- cycle ;

\draw    (80,180) .. controls (189.95,115.32) and (465.72,100.15) .. (571.72,166.99) ;
\draw [shift={(573.3,168)}, rotate = 212.98] [fill={rgb, 255:red, 0; green, 0; blue, 0 }  ][line width=0.75]  [draw opacity=0] (8.93,-4.29) -- (0,0) -- (8.93,4.29) -- cycle    ;

\draw   (220,160) -- (314.3,160) -- (314.3,200) -- (220,200) -- cycle ;

\draw   (365.7,210) -- (460,210) -- (460,250) -- (365.7,250) -- cycle ;

\draw   (535.7,180) -- (630,180) -- (630,220) -- (535.7,220) -- cycle ;

\draw   (220,210) -- (314.3,210) -- (314.3,250) -- (220,250) -- cycle ;

\draw    (150,210) -- (198.14,229.26) ;
\draw [shift={(200,230)}, rotate = 201.8] [fill={rgb, 255:red, 0; green, 0; blue, 0 }  ][line width=0.75]  [draw opacity=0] (8.93,-4.29) -- (0,0) -- (8.93,4.29) -- cycle    ;

\draw    (198.14,180.74) -- (150,200) ;

\draw [shift={(200,180)}, rotate = 158.2] [fill={rgb, 255:red, 0; green, 0; blue, 0 }  ][line width=0.75]  [draw opacity=0] (8.93,-4.29) -- (0,0) -- (8.93,4.29) -- cycle    ;

\draw (410.5,177) node  [align=left] {$\displaystyle s_{t, critic}$};
\draw (499.8,229) node  [align=left] {$\displaystyle a_{t}$};
\draw (505,265) node  [align=left] {$\displaystyle A$};
\draw (78.85,210) node   {$Environment$};
\draw (162,181) node  [align=left] {$\displaystyle s_{t}$};
\draw (318,111) node  [align=left] {$\displaystyle r_{t}$};
\draw (412.85,230) node   {$Actor$};
\draw (267.15,180) node   {$Attention$};
\draw (582.85,200) node   {$Critic$};
\draw (267.15,230) node   {$Attention$};
\draw (163,224) node  [align=left] {$\displaystyle s_{t}$};
\draw (340,214) node  [align=left] {$\displaystyle s_{t, actor}$};

\end{tikzpicture}
    \caption{The \acrshort*{atta2c} architecture}
    \label{fig:atta2c_arch}
\end{figure}

Taking this similarity a step further, we conjecture that separating the feature space into two parts can be advantageous, as different features may be useful for the actor and for the critic. The proposed architecture aims to overcome this disadvantage by utilizing attention. As shown in \autoref{fig:atta2c_arch}, the input from the environment is fed through separate attention layers, thus both heads can put emphasis onto the most important parts of the latent process, which is formulated as follows:
\begin{align}
    \at &= Actor\parenthesis{Attn_{\gls*{policy}}\parenthesis{\phit}} \\
    \advt &= Critic\parenthesis{Attn_{\gls*{advantage}}\parenthesis{\phit}},
\end{align}
where $Actor$ (which is basically the policy \gls*{policy}) and $Critic$ denote the corresponding networks, \gls*{feat} is the feature transform of the input (if present), while $Attn$ is a shorthand for the probability distribution of the attention mechanism. Subscripts denote that two separate attention layers are used to predict the policy \gls*{policy} and the advantage function \gls*{advantage}.

\begin{table}\centering
    \begin{tabular}{@{}ll@{}}
    	\toprule
    	\textbf{\acrshort*{gan}} & \textbf{\acrlong*{ac}} \\ \midrule
    	Generator          & Actor             \\
    	Discriminator      & Critic            \\
    	Label              & Reward            \\
    	Input              & \textbf{State}    \\
    	Latent space       & \textbf{State}    \\ \bottomrule
    \end{tabular}
    \caption{The connection between \gls*{gan} and \gls*{ac} concepts~\cite{ac_gan}}
    \label{tab:ac_gan}
\end{table}

\subsection{Feature- and action selective ICM}
\label{subsec:att_icm}
Being able to separate the most useful features cannot only be advantageous for the \acrlong*{ac} network, but also for the curiosity formulation. Thus, we introduce attention mechanism to \gls*{icm} to make it action- and feature-selective. To do that, attention is applied onto the concatenation of the inputs of the forward and inverse models (denoted by the $fwd$ and $inv$ subscripts, respectively):
\begin{align}
    \hat{\gls*{feat}}_{t+1} &= Forward\parenthesis{Attn_{fwd}\parenthesis{\brackets{\phit, \at}}} \\ 
    \hat{\gls*{action}}_{t} &= Inverse\parenthesis{Attn_{inv}\parenthesis{\brackets{\phit, \at}}}, 
\end{align}
where $\hat{\cdot}$ stands for predicted values. The equations above describe the single attention case, but we also experimented with double attention, in which we swapped the order of concatenation and attention. In this case, the attention-weighted features and actions are concatenated. The reason for this second formulation was to separate the weighting between the two domains. The main advantage of the latter formulation could be that using double attention could ensure that both in the feature and action spaces there will be a subspace which is emphasized. This is not the case when using single attention, which can be problematic if the distributions are not in the same order of magnitude, implying that one domain is more important due to using different value scales.

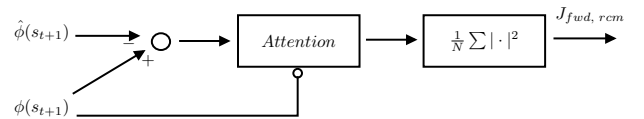
\begin{figure}[b]
    \centering
    \begin{tikzpicture}[x=0.75pt,y=0.75pt,yscale=-1,xscale=1, scale=0.65, every node/.style={scale=0.65}]

\draw    (147,93) -- (185,93) ;
\draw [shift={(187,93)}, rotate = 180] [fill={rgb, 255:red, 0; green, 0; blue, 0 }  ][line width=0.75]  [draw opacity=0] (8.93,-4.29) -- (0,0) -- (8.93,4.29) -- cycle    ;

\draw    (67.5,151) -- (239,151) -- (239,120.35) ;
\draw [shift={(239,118)}, rotate = 270] [color={rgb, 255:red, 0; green, 0; blue, 0 }  ][line width=0.75]      (0, 0) circle [x radius= 3.35, y radius= 3.35]   ;

\draw   (192.7,73) -- (287,73) -- (287,113) -- (192.7,113) -- cycle ;

\draw    (119.29,101.03) -- (64.5,134) ;

\draw [shift={(121,100)}, rotate = 148.96] [fill={rgb, 255:red, 0; green, 0; blue, 0 }  ][line width=0.75]  [draw opacity=0] (8.93,-4.29) -- (0,0) -- (8.93,4.29) -- cycle    ;
\draw    (66.5,89) -- (115,89) ;
\draw [shift={(117,89)}, rotate = 180] [fill={rgb, 255:red, 0; green, 0; blue, 0 }  ][line width=0.75]  [draw opacity=0] (8.93,-4.29) -- (0,0) -- (8.93,4.29) -- cycle    ;

\draw   (125.85,93.5) .. controls (125.85,89.91) and (129.02,87) .. (132.93,87) .. controls (136.83,87) and (140,89.91) .. (140,93.5) .. controls (140,97.09) and (136.83,100) .. (132.93,100) .. controls (129.02,100) and (125.85,97.09) .. (125.85,93.5) -- cycle ;
\draw   (336.7,73) -- (431,73) -- (431,113) -- (336.7,113) -- cycle ;

\draw    (292,92) -- (310.5,92) -- (330,92) ;
\draw [shift={(332,92)}, rotate = 180] [fill={rgb, 255:red, 0; green, 0; blue, 0 }  ][line width=0.75]  [draw opacity=0] (8.93,-4.29) -- (0,0) -- (8.93,4.29) -- cycle    ;

\draw    (437.5,91) -- (456,91) -- (484.5,91) ;
\draw [shift={(486.5,91)}, rotate = 180] [fill={rgb, 255:red, 0; green, 0; blue, 0 }  ][line width=0.75]  [draw opacity=0] (8.93,-4.29) -- (0,0) -- (8.93,4.29) -- cycle    ;

\draw (239.85,93) node   {$Attention$};
\draw (40.5,145) node  [align=left] {$\displaystyle \phi ( s_{t+1})$};
\draw (40.5,85) node  [align=left] {$\displaystyle \hat{\phi }( s_{t+1})$};
\draw (466,73) node  [align=left] {$\displaystyle J_{fwd,\ rcm}$};
\draw (383.85,93) node   {$\frac{1}{N}\sum |\cdot |^{2}$};
\draw (108,95) node  [align=left] {$\displaystyle -$};
\draw (123,108) node  [align=left] {$\displaystyle +$};

\end{tikzpicture}
    \caption{Forward loss formulation of the \acrshort*{rcm} agent}
    \label{fig:rcm_fwd_loss}
\end{figure}

\subsection{Rational curiosity}
\label{subsec:rational_cur}

The curiosity-driven exploration strategies of ~\cite{icm} and ~\cite{disagreement} represent a rather powerful approach to train agents in sparse reward settings. Nonetheless, these models are based on the assumption that enforcing curiosity-driven exploration is a good choice for every state and action, even though this is not true in every scenario. \Eg consider the noisy TV experiment of ~\cite{large_scale_curiosity}, where the agent has control over generating new (and thus, unexpected) instances of random noise, resulting in high values of the \gls*{icm}/disagreement losses.

To classify the states based on the usefulness of curiosity, the state space \stsp can be divided into two subsets: \stsp[+] and \stsp[-], where the former denotes the subset of states with useful curiosity, while the latter the useless or harmful curiosity settings. Curiosity is termed as harmful if being driven by this curiosity does not ensure the fulfillment of the original objective, not even in the long run. \Ie sacrificing short-term rewards to develop general skills is useful, but overfitting random noise is not. 

Thus, the rational objective of curiosity-driven exploration would be to minimize the probability of being in states in \stsp[-], which corresponds to:
\begin{align}
    min \ \; \gls*{state_trans_prob}\parenthesis{\st[t+1] \in \stsp[-] | \st, \at}, \label{eq:rat_cur_obj}
\end{align}
\ie it is intended to select actions which do not lead the agent into \stsp[-], independent from its actual state.

\begin{figure}[tb]
    \centering 
    \begin{tikzpicture}[x=0.75pt,y=0.75pt,yscale=-1,xscale=1, scale=0.55, every node/.style={scale=0.55}]

\draw   (77.3,50) -- (77.3,144.3) -- (52.3,144.3) -- (52.3,50) -- cycle ;

\draw  [dash pattern={on 0.84pt off 2.51pt}] (40,34.3) -- (270,34.3) -- (270,160) -- (40,160) -- cycle ;
\draw   (112.3,50) -- (112.3,144.3) -- (87.3,144.3) -- (87.3,50) -- cycle ;

\draw   (147.3,50) -- (147.3,144.3) -- (122.3,144.3) -- (122.3,50) -- cycle ;

\draw   (182.3,50) -- (182.3,144.3) -- (157.3,144.3) -- (157.3,50) -- cycle ;

\draw   (222.3,50) -- (222.3,144.3) -- (197.3,144.3) -- (197.3,50) -- cycle ;

\draw   (260,50) -- (260,144.3) -- (235,144.3) -- (235,50) -- cycle ;

\draw  [fill={rgb, 255:red, 222; green, 222; blue, 222 }  ,fill opacity=1 ] (329.4,38) -- (423.7,38) -- (423.7,78) -- (329.4,78) -- cycle ;

\draw  [fill={rgb, 255:red, 222; green, 222; blue, 222 }  ,fill opacity=1 ] (329.4,108) -- (423.7,108) -- (423.7,148) -- (329.4,148) -- cycle ;

\draw   (501.7,54.2) -- (596,54.2) -- (596,94.2) -- (501.7,94.2) -- cycle ;

\draw   (501.7,102.2) -- (596,102.2) -- (596,142.2) -- (501.7,142.2) -- cycle ;

\draw  [dash pattern={on 0.84pt off 2.51pt}] (489.4,43.2) -- (609,43.2) -- (609,150) -- (489.4,150) -- cycle ;

\draw    (279.4,58) -- (317.4,58) ;
\draw [shift={(319.4,58)}, rotate = 180] [fill={rgb, 255:red, 0; green, 0; blue, 0 }  ][line width=0.75]  [draw opacity=0] (8.93,-4.29) -- (0,0) -- (8.93,4.29) -- cycle    ;

\draw    (279.4,128) -- (317.4,128) ;
\draw [shift={(319.4,128)}, rotate = 180] [fill={rgb, 255:red, 0; green, 0; blue, 0 }  ][line width=0.75]  [draw opacity=0] (8.93,-4.29) -- (0,0) -- (8.93,4.29) -- cycle    ;

\draw    (439.4,58) -- (477.4,58) ;
\draw [shift={(479.4,58)}, rotate = 180] [fill={rgb, 255:red, 0; green, 0; blue, 0 }  ][line width=0.75]  [draw opacity=0] (8.93,-4.29) -- (0,0) -- (8.93,4.29) -- cycle    ;

\draw    (439.4,128) -- (477.4,128) ;
\draw [shift={(479.4,128)}, rotate = 180] [fill={rgb, 255:red, 0; green, 0; blue, 0 }  ][line width=0.75]  [draw opacity=0] (8.93,-4.29) -- (0,0) -- (8.93,4.29) -- cycle    ;

\draw (209.8,97.15) node [rotate=-90]  {$AvgPool2d$};
\draw (169.8,97.15) node [rotate=-90]  {$Conv2d$};
\draw (134.8,97.15) node [rotate=-90]  {$Conv2d$};
\draw (99.8,97.15) node [rotate=-90]  {$Conv2d$};
\draw (64.8,97.15) node [rotate=-90]  {$Conv2d$};
\draw (247.5,97.15) node [rotate=-90]  {$LSTM$};
\draw (506.2,26.5) node [scale=0.8]  {$AC$};
\draw (548.85,122.2) node   {$Critic$};
\draw (548.85,74.2) node   {$Actor$};
\draw (376.55,128) node   {$Attention$};
\draw (376.55,58) node   {$Attention$};
\draw (90.3,18.5) node [scale=0.8]  {$FeatureExtractor$};

\end{tikzpicture}
	\caption{The \gls*{a2c}/\gls*{atta2c} network (gray indicates optional layer)}
	\label{fig:ac_nn}
\end{figure}
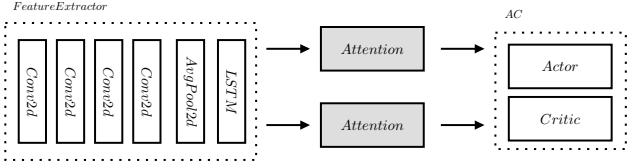

\begin{figure}[tb]
    \centering 
    \begin{tikzpicture}[x=0.75pt,y=0.75pt,yscale=-1,xscale=1, scale=0.55, every node/.style={scale=0.55}]

\draw   (67.3,80) -- (67.3,174.3) -- (42.3,174.3) -- (42.3,80) -- cycle ;

\draw  [dash pattern={on 0.84pt off 2.51pt}] (30,64.3) -- (232.3,64.3) -- (232.3,190) -- (30,190) -- cycle ;
\draw   (102.3,80) -- (102.3,174.3) -- (77.3,174.3) -- (77.3,80) -- cycle ;

\draw   (137.3,80) -- (137.3,174.3) -- (112.3,174.3) -- (112.3,80) -- cycle ;

\draw   (172.3,80) -- (172.3,174.3) -- (147.3,174.3) -- (147.3,80) -- cycle ;

\draw   (212.3,80) -- (212.3,174.3) -- (187.3,174.3) -- (187.3,80) -- cycle ;

\draw  [fill={rgb, 255:red, 222; green, 222; blue, 222 }  ,fill opacity=1 ] (315,85) -- (409.3,85) -- (409.3,125) -- (315,125) -- cycle ;

\draw    (240,107) -- (278,107) ;
\draw [shift={(280,107)}, rotate = 180] [fill={rgb, 255:red, 0; green, 0; blue, 0 }  ][line width=0.75]  [draw opacity=0] (8.93,-4.29) -- (0,0) -- (8.93,4.29) -- cycle    ;

\draw    (419,107) -- (457,107) ;
\draw [shift={(459,107)}, rotate = 180] [fill={rgb, 255:red, 0; green, 0; blue, 0 }  ][line width=0.75]  [draw opacity=0] (8.93,-4.29) -- (0,0) -- (8.93,4.29) -- cycle    ;

\draw  [fill={rgb, 255:red, 222; green, 222; blue, 222 }  ,fill opacity=1 ] (315,134) -- (409.3,134) -- (409.3,174) -- (315,174) -- cycle ;

\draw    (240,153) -- (278,153) ;
\draw [shift={(280,153)}, rotate = 180] [fill={rgb, 255:red, 0; green, 0; blue, 0 }  ][line width=0.75]  [draw opacity=0] (8.93,-4.29) -- (0,0) -- (8.93,4.29) -- cycle    ;

\draw    (419,153) -- (457,153) ;
\draw [shift={(459,153)}, rotate = 180] [fill={rgb, 255:red, 0; green, 0; blue, 0 }  ][line width=0.75]  [draw opacity=0] (8.93,-4.29) -- (0,0) -- (8.93,4.29) -- cycle    ;

\draw   (462.3,86.2) -- (556.6,86.2) -- (556.6,126.2) -- (462.3,126.2) -- cycle ;

\draw   (462.3,134.2) -- (556.6,134.2) -- (556.6,174.2) -- (462.3,174.2) -- cycle ;

\draw  [dash pattern={on 0.84pt off 2.51pt}] (300,75.2) -- (569.6,75.2) -- (569.6,182) -- (300,182) -- cycle ;

\draw (54.8,127.15) node [rotate=-90]  {$Conv2d$};
\draw (89.8,127.15) node [rotate=-90]  {$Conv2d$};
\draw (124.8,127.15) node [rotate=-90]  {$Conv2d$};
\draw (159.8,127.15) node [rotate=-90]  {$Conv2d$};
\draw (199.8,127.15) node [rotate=-90]  {$AvgPool2d$};
\draw (80.3,48.5) node [scale=0.8]  {$FeatureExtractor$};
\draw (362.15,105) node   {$Attention$};
\draw (362.15,154) node   {$Attention$};
\draw (509.45,154.2) node   {$Inverse$};
\draw (509.45,106.2) node   {$Forward$};
\draw (315.5,58.5) node [scale=0.8]  {$ICM$};

\end{tikzpicture}
	\caption{The \gls*{icm}/\gls*{rcm} network (gray indicates optional layer)}
	\label{fig:icm_nn}
\end{figure}
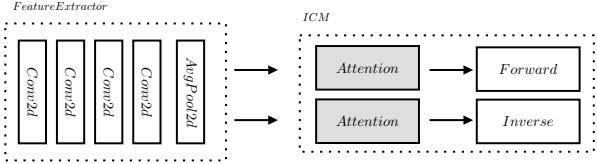

To achieve that, an attention network~\cite{attention} is used on the forward loss function of the \gls*{icm} module~\cite{icm} - the inverse loss is not used, as for each action prediction, there is only one true action, where weighting has no effect. The new forward loss term is illustrated in \autoref{fig:rcm_fwd_loss}, where the round-headed arrow denotes that \phit[t+1] is used as the control term in the attention layer, \ie it determines the weighting of the \gls*{mse} (which is here decomposed into two steps, \ie subtraction and squared mean) through a probability distribution. Thus, the loss function is modified in the following manner:
\begin{align}
    \gls*{cost_rcm} = \gls*{cost_a2c} + \gls*{icm_beta}Attn\parenthesis{\gls*{cost_fwd}, \phit[t+1]} + \parenthesis{1-\gls*{icm_beta}}\gls*{cost_inv},
\end{align}
where \gls*{rcm} is the term for the new model,  \gls*{cost_a2c} is the objective function for the \gls*{a2c} network, \gls*{cost_fwd} for the forward and \gls*{cost_inv} for the inverse dynamics, weighted with a scalar \gls*{icm_beta}. This formula is motivated by the fact that \phit[t+1] encodes fully whether whether the state of the agent is in \stsp[+] or in \stsp[-]. We hypothesize that this formulation can help to utilize curiosity only in situations where the agent can benefit from that on the long run, but omits curiosity otherwise.

\subsection{Implementation}
\label{subsec:implementation}
The proposed methods are implemented in PyTorch~\cite{Paszke2017}, the agents are based on the implementation of~\cite{icm} (shown in \autoref{fig:ac_nn} and \autoref{fig:icm_nn}). The extension is indicated with the gray boxes, as those layers are part of some of the proposed methods discussed in the following. Five configurations were implemented: \gls*{atta2c}, a single- and double-attention \gls*{icm} and \gls*{rcm}, and the traditional \gls*{icm} agent is used as a baseline~\cite{icm}.

For test purposes, the Atari environments of the OpenAI Gym~\cite{gym} are used via the  Stable-baselines~\cite{stable-baselines} package. Three environments were chosen, \ie Breakout, Pong and Seaquest (as in~\cite{disagreement}), which provide single- (Pong, Seaquest) and multiplayer (Pong) tasks with one (Breakout, Pong) and multiple (Seaquest) objectives. For each of the three, both the deterministic (v4) and the stochastic (v0, with nonzero action repeat probability) variants were evaluated. All agents were trained on 4 environments in a parallel manner for 2,500,000 rollouts with 5 steps each, using 4 stacked frames.

\begin{figure}[tb]
    \centering
    \includegraphics[width=.7\linewidth,height=.7\textheight,keepaspectratio]{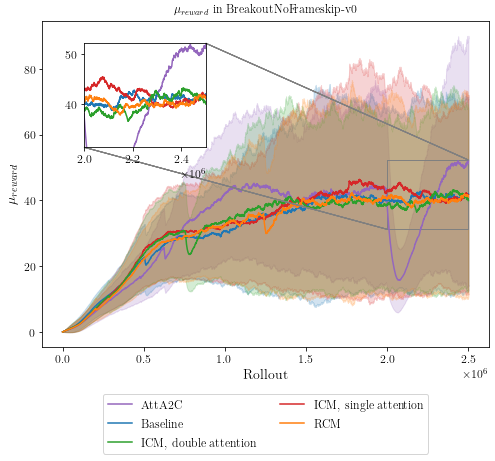}
	\caption{Mean reward in the Breakout environment (v0)}
	\label{fig:Breakout_v0_rwd}
\end{figure}

\begin{figure}[htb]
    \centering
    \includegraphics[width=.7\linewidth,height=.7\textheight,keepaspectratio]{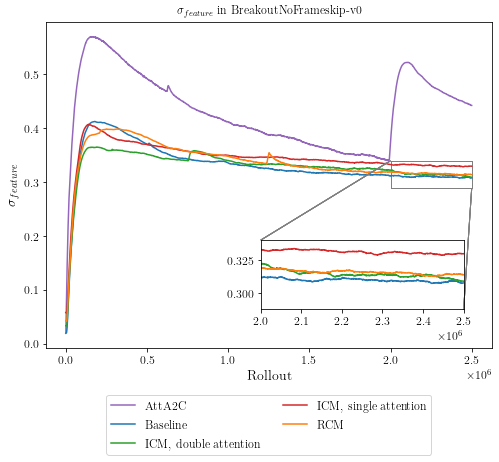}
	\caption{Feature standard deviation in the Breakout environment (v0)}
	\label{fig:breakout_v0_feat}
\end{figure}

\section{Results}
\label{sec:results}
We monitored two metrics to compare both performance and generalization: we used the mean reward for the former, and the mean standard deviation of the features for the latter. Due to space restrictions, only the results for the more difficult, stochastic v0 environments are depicted here, but the others are available in the GitHub repository.

\begin{figure}[htb]
    \centering
    \includegraphics[width=.7\linewidth,height=.7\textheight,keepaspectratio]{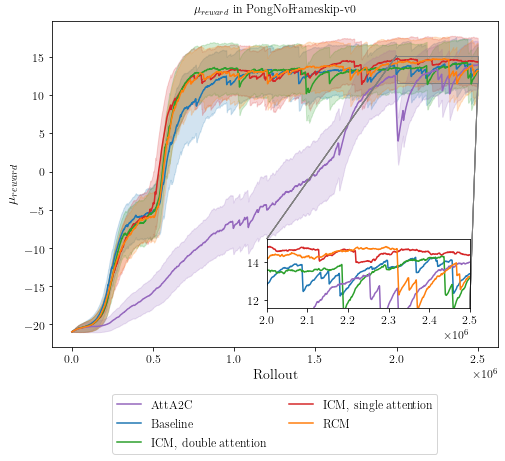}
	\caption{Mean reward in the Pong environment (v0)}
	\label{fig:pong_v0_rwd}
\end{figure}

\begin{figure}[htb]
    \centering
    \includegraphics[width=.7\linewidth,height=.7\textheight,keepaspectratio]{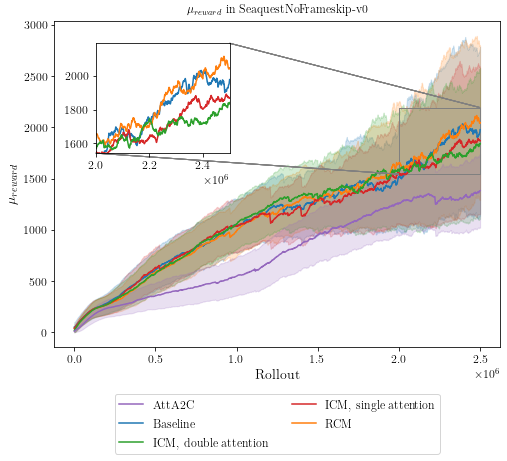}
	\caption{Mean reward in the Seaquest environment (v0)}
	\label{fig:seaquest_v0_rwd}
\end{figure}

For Breakout (\autoref{fig:Breakout_v0_rwd}, showing a $1 \sigma$ confidence interval as well),  the selective versions of \gls*{icm} had a consistently good performance, but the \gls*{atta2c} agent trained faster at the beginning, and after experiencing a jump in variance, it managed to significantly outperform the other agents. The standard deviation of the feature space shown in \autoref{fig:breakout_v0_feat} visualizes the general setting, \ie the significantly higher values in case of \gls*{atta2c} (for every environment, but the "jumps" are smaller). In our experiments, agents (not concerning \gls*{atta2c} for this statement)  with higher standard deviation performed generally better. In case of Pong (\autoref{fig:pong_v0_rwd}), the single-attention \gls*{icm} performed as the best, followed by the \gls*{rcm} agent. In this case, \gls*{atta2c} trained slower, but managed to achieve comparable rewards - the reason for this could be the smaller gradients due to attention between feature space and the actor/critic. The \gls*{rcm} agent was the best for the most complex environment, Seaquest, as \autoref{fig:seaquest_v0_rwd} shows. In this case, the selective \gls*{icm} agents were overtaken by the original \gls*{icm}, while \gls*{atta2c} experienced much slower training. 

\begin{table}\centering
	\begin{tabular}{@{}lrr@{}}
	\toprule
		\textbf{Agent}                 &      \textbf{Mean normalized reward} &\textbf{Std. dev.} \\
	\midrule
		Baseline                       &           $92.33$ &$4.98$ \\
		\acrshort*{atta2c}              &           $88.01$ &$13.36$ \\
		$\acrshort*{icm}_{1\times Att}$ &           $94.88$ &$\mathbf{4.46}$ \\
		$\acrshort*{icm}_{2\times Att}$ &          $92.17$& $5.68$ \\
		\acrshort*{rcm}                 & $\mathbf{95.31}$ &$6.48$ \\
	\bottomrule
	\end{tabular}
	\caption{Comparison of the agents' normalized mean performance (higher is better)}
	\label{tab:agent_perf}
\end{table}

To be able to provide a concise but quantitative summary of the agents' performance in all environments (including both variants), we normalized the highest rewards of each agent, with 100 denoting the best performance. This way, we were able to compare the mean performances in a relative manner, summarized in \autoref{tab:agent_perf} (including a $1 \sigma$ confidence interval). As it shows, the \gls*{rcm} agent performed the best, followed by the single attention \gls*{icm}. Note that \gls*{atta2c} had both the lowest mean and the highest variance, mainly due to the good performance in Breakout, but a moderate one in the other scenarios.

\section{Discussion}
\label{sec:discussion}
This work investigated the paradigm of curiosity-driven exploration in \gls*{rl}, which was extended by the attention mechanism. We proposed three different methods to incorporate attention for utilizing curiosity in a selective manner. The new models were tested in the OpenAI Gym environment and have shown consistent improvement to the baseline models used for comparison.


\vfill
\pagebreak
\bibliographystyle{IEEEbib}
\bibliography{Template}

\end{document}